\ifdefined\pdfoutput\pdfoutput=1\fi 
\documentclass[11pt]{article}
\usepackage[preprint]{acl}

\usepackage{times}
\usepackage{latexsym}
\usepackage[T1]{fontenc}
\usepackage[utf8]{inputenc}
\usepackage{microtype}
\usepackage{inconsolata}
\usepackage{graphicx}
\usepackage{subcaption}
\usepackage{booktabs}
\usepackage{multirow}
\usepackage{float}
\usepackage{verbatim}
\usepackage[most]{tcolorbox}
\usepackage{xcolor}
\definecolor{pastelFrame}{RGB}{173,216,230} 
\definecolor{pastelTitle}{RGB}{213,234,249} 
\definecolor{pastelBack}{RGB}{250,250,250}  

\DeclareUnicodeCharacter{2212}{-}               
\DeclareUnicodeCharacter{2192}{\textrightarrow} 
\DeclareUnicodeCharacter{2013}{--}              
\DeclareUnicodeCharacter{2014}{---}             
\DeclareUnicodeCharacter{2248}{\approx}         

\usepackage[section]{placeins}
\title{The Brittle Compass: Navigating LLM Prompt Sensitivity\\ in Slovak Migration Media Discourse%
  \thanks{Published in the Proceedings of the First Workshop on Advancing NLP for Low-Resource Languages
  (LowResNLP 2025), associated with RANLP 2025, Varna, Bulgaria, pages 88--101,
  \url{https://doi.org/10.26615/978-954-452-100-4-010}.}}

\author{Samuel Harvan \\
  Central European University,\\ Vienna, Austria \\
  Kempelen Institute \\of Intelligent Technologies, \\Bratislava, Slovakia \\
  \texttt{samuel.harvan@intern.kinit.sk} \\\And
  Jaroslav Kopčan \\
  Kempelen Institute \\of Intelligent Technologies,\\ Bratislava, Slovakia \\
  \texttt{jaroslav.kopcan@kinit.sk} \\\AND
  Marek Šuppa \\
  Cisco Systems \\
  Comenius University \\in Bratislava, Slovakia \\
  \texttt{marek@suppa.sk} \\\And
  Andrej Findor \\
  Comenius University \\in Bratislava, Slovakia \\
  \texttt{andrej@findor.fses.uniba.sk} \\}

\begin{document}
\maketitle
\begin{abstract}

In this work, we present a case study that explores various tasks centered around the topic of migration in Slovak, a low-resource language, such as topic relevance and geographical relevance classification, and migration source/destination location term extraction. Our results demonstrate that native (Slovak) prompts yield a modest, task-dependent gains, while large models show significant robustness to prompt variations compared to their smaller counterparts. Analysis reveals that instructions (system or task) emerge as the most critical prompt component, more so than the examples sections, with task-specific performance benefits being more pronounced than overall language effects.

\end{abstract}

\section{Introduction}


Large Language Models (LLMs) have become a cornerstone in addressing a multitude of Natural Language Processing (NLP) tasks, significantly transforming the field by enabling machines to understand and generate human-like text. Their prominence is particularly notable in multilingual contexts, driven by their strong zero-shot and few-shot performance, especially when coupled with sophisticated reasoning mechanisms \cite{vatsal2025multilingualpromptengineeringlarge}. These models, often trained on vast datasets, exhibit proficiency across a diverse array of languages and have demonstrated effectiveness in numerous downstream applications, including tasks such as natural language understanding, common-sense reasoning, and question-answering, thereby capturing both the syntax and semantics of texts \cite{vatsal2025multilingualpromptengineeringlarge}. Modern multilingual LLMs are capable of performing tasks across more than 100 languages, representing a significant breakthrough in NLP \cite{vatsal2025multilingualpromptengineeringlarge}.

The typical mode of operation for these models involves prompt- or context-engineering, where specific instructions are provided to guide the LLM towards correctly solving the task at hand \cite{wahle2024paraphrasetypes,lu2024promptsdifferenttermssensitivity}. However, the ultimate efficacy of this approach is heavily contingent upon the nuances of the prompt employed, including its formatting and the organization of data within it \cite{he2024doespromptformattingimpact,MOF-ngweta2025llmsrobustnesschangesprompt,japtext-gan2023sensitivityrobustnesslargelanguage,razavi2025benchmarkingpromptsensitivitylarge}. This sensitivity is of particular importance in multilingual scenarios. The language in which a prompt induces an LLM to perform the reasoning component of its computation can exert a significant influence on the final performance \cite{poelman2024rolesenglish}. For instance, LLMs may struggle to adhere to all specified rules within complex prompts, and innovative prompting strategies, such as translating error-prone rules into different languages, have been proposed to enhance their reasoning and understanding \cite{wang2025mlprompt}. Research is actively exploring methods to improve multilingual reasoning, with a focus on augmenting the ability of LLMs to handle diverse languages and intricate reasoning tasks \cite{vatsal2025multilingualsurvey}. Techniques like multilingual instruction tuning and dynamic, language-aware prompting (e.g., language-specific trigger tokens) aim to bolster reasoning capabilities and consistency across various languages \cite{polyprompt-roll2025polypromptautomatingknowledgeextraction,vatsal2025multilingualpromptengineeringlarge}.

Despite these advancements, challenges persist, particularly for low-resource languages, which often suffer from a scarcity of training data and computational resources. Cross-lingual transfer learning, which leverages data and models from high-resource languages, is a key research area for improving NLP performance in these settings \cite{vatsal2025multilingualpromptengineeringlarge}. Slovak, for example, is a language where dedicated transformer-based models like SlovakBERT have been developed to establish benchmarks and advance NLP capabilities \cite{pikuliak2022slovakbertslovakmaskedlanguage}. The investigation of abstract multilingual reasoning, especially for extremely low-resource languages, often involves inducing linguistic patterns from seed exemplars through methods like analogical prompting \cite{vatsal2025multilingualpromptengineeringlarge}.

In this work, we present a case study that explores the application of LLMs to various tasks centered around the topic of migration in Slovak, a low-resource language. Specifically, we explore tasks such as topic relevance classification, geographical relevance classification, and the extraction of migration source and destination location terms. This study aims to illuminate the intricacies of prompting LLMs for specialized tasks in a low-resource linguistic context, with a particular focus on how prompt design and reasoning language affect performance in migration-related NLP applications.

\section{Related Work}
The effectiveness of Large Language Models (LLMs) in multilingual contexts has become a critical research area, particularly as these models demonstrate varying performance across different languages and cultural contexts.
The field of multilingual prompt engineering has emerged as a crucial technique for enhancing LLM performance across diverse linguistic landscapes. Comprehensive overviews usually confirm significant disparities in research attention, with high-resource languages getting substantially more focus than their low-resource counterparts, “other languages”. While most NLP tasks are heavily concentrated in high-resource language settings, there is motivation to bridge these domains through cross-lingual transfer learning \cite{vatsal2025multilingualpromptengineeringlarge}. An important fact is that multilingual prompt engineering faces unique challenges to ensure consistent performance across languages, as LLMs often exhibit disparities in performance depending on the availability of training data for different languages. This finding directly relates to our focus on Slovak as a low-resource language, where such disparities become limiting in specialized domains. 

The sensitivity of LLMs to prompt formulation and formatting has been identified as a critical factor affecting model performance. The concept of prompt sensitivity prediction demonstrates that small variations in prompt phrasing, structure, or punctuation can lead to substantially different outputs, even totally misleading the LLMs on tasks they previously solved correctly. \cite{razavi2025benchmarkingpromptsensitivitylarge}. This serves as the foundational work that formalizes prompt sensitivity as a systematic challenge when working with LLMs. Moreover, systematic examination of the impact of different prompt templates on LLM performance across various tasks results in performance variation. Different template selections can cause the performance to fluctuate by up to 40\% on smaller LLMs, while larger ones demonstrate greater robustness to these format variations \cite{he2024doespromptformattingimpact}. These findings suggest that prompt formatting considerations become more critical when working with low-resource languages; however, there is no universally optimal format across the usual NLP tasks. The effectiveness of each is highly context-dependent on models, tasks, or context window sizes. 
Inspection of prompt sensitivity by examining how different prompt components interact with model architectures could provide insights into why sensitivity occurs. For instance, CoT prompting significantly increases sensitivity to variations despite maintaining similar accuracy in comparison to basic ‘static’ prompts. \cite{lu2024promptsdifferenttermssensitivity}. In general, instructions seem to provide more stable performance than complex ongoing reasoning. The Mixture of Format (MoF) addresses the prompt brittleness problem by deliberately varying the formatting of few-shot examples rather than seeking a single one-size-fits-all optimal format. MoF maintains semantic content while diversifying the textual format. This results in improved robustness compared to traditional fixed-format prompts while preserving task performance. The approach aims to reduce prompt brittleness across various LLMs and tasks \cite{MOF-ngweta2025llmsrobustnesschangesprompt}.

The reasoning mechanisms in multilingual settings have received a considerable amount of attention, particularly regarding Chain-of-Thought (CoT) prompting strategies, in order to remedy performance disparities across languages. Methods like XLT \cite{XLT-huang-etal-2023-languages} demonstrate a systematic approach where LLMs first translate the input from the native language to English, solve the task, generate reasoning chains, and then format the output accordingly, while consistently outperforming other approaches across multiple datasets.
Building on this concept, \textit{Cross-lingual Prompting} (CLP) \cite{CLP-qin2023crosslingualpromptingimprovingzeroshot} introduces a two-step process focusing on cross-lingual alignment, where the model generates reasoning chains in English rather than the native language to establish representations between languages. Interestingly, the whole process could be extended by introducing greater linguistic flexibility - \textit{Cross-lingual self-consistent prompting} (CLSP), allowing LLMs to comprehend tasks and employ reasoning steps in different languages before selecting the most consistent answer across different language-based reasoning chains, implying that English is not always the best default option.
The same could be inferred from the evaluation of language-specific optimization, represented as a parameter-efficient framework that learns language-specific trigger tokens through gradient-based search \cite{polyprompt-roll2025polypromptautomatingknowledgeextraction}. Results show that autoprompting like this can yield significant performance improvements over static, translated prompts. Even without the autogeneration, the language-specific prompt engineering can be effective with systematic prompt template adaptation for specific languages \cite{japtext-gan2023sensitivityrobustnesslargelanguage}


\section{Dataset}
For experimental evaluations, we employ a theme-specific Slovak annotated dataset that classifies content for multiple tasks. This dataset focuses on analyzing how migration is portrayed in Slovak media from 2003 to 2022, by examining individual media pieces - news articles. The key classification dimensions are:

\begin{itemize}
    \item \textbf{Theme Relevance} on article-level is about categorizing content according to its connection to human migration within the specified timeframe, with classifications of \textit{strong}, \textit{weak}, or \textit{not-relevant}.
    \item \textbf{Geographical Relevance} on sentence-level distinguishes between content that pertains to Slovakia (i.e., migration \textit{to}, \textit{from}, or \textit{through} the country), versus content that does not relate to the country.
    \item \textbf{Location Extraction} on sentence-level facilitates an extraction task, with sentences annotated by identified source and target locations, according to the annotation guidelines.
\end{itemize}

For \textit{Theme Relevance}, we have used a subset from the whole corpus, given its extensive volume (1,800,000 articles from years 2003-2024). The subset was created by stratified sampling, which was applied annually. Every article item in the subset received annotations from a minimum of three separate annotators using an Argilla interface, and only those instances where the majority of annotators agreed were retained in the final dataset. Inter-annotator agreement, measured by Krippendorff’s $\alpha$, was 0.326, indicating only low-moderate agreement; we note this as a limitation and encourage cautious interpretation. For more details on the original dataset, see \cite{hamerlik-etal-2024-chatgpt} and the Appendix \ref{sec:theme_data_statistic}

For the  \textit{Locality Extraction} and \textit{Geographical Relevance} tasks, we have manually curated a dataset comprising several thousand sentences on migration, sourced from original Slovak-media articles published in 2022 and 2024, as a subset. This dataset is therefore partitioned into two subsets tailored for the aforementioned tasks. While many sentences overlap between subsets, some are exclusive due to task-specific relevance. The sentences cover migration related to conflicts in Ukraine, Syria, and Gaza, supplemented by other diverse scenarios (e.g., political or economic migration) to ensure broad representation. The annotation focused on identifying source and target migration locations, excluding purely transit mentions. Near-identical sentences derived from modified press releases were deduplicated.
During the annotations, three authors conducted manual annotation following the guidelines detailed in Appendix \ref{sec:guidelines}; while sentences lacking complete annotator agreement were removed to maintain data quality. For more information about the datasets see the Appendix section \ref{sec:loc_geo_data_statistic} 

The dataset comprises a thorough compilation of human-labeled sentences focused on migration topics, sourced from 2323 distinct articles. Two specialized subsets were created from this collection: a Slovakia-focused subset with 2736 annotated examples, and a geographic locality extraction subset containing 1652 human-annotated samples designed for identifying and extracting location information. The complete dataset was divided using stratified sampling with a 70:20:10 distribution for training, validation, and testing sets, maintaining balanced class representation across all partitions.

\section{Experiments}
The following section represents the results from non-reasoning as well as reasoning model within the prompt sensibility case study in native and english ‘default’ language setup.

Based on the results outlined in Table \ref{tab:experiments_matrix} on non-reasoning and reasoning model testing, there are some insights to be inferred relevant to the prompt strategy. 
The main setup for experiments with reasoning models was about forcing them to reason in their native language. However we only managed to force two models to reason in Slovak: \textit{grok-3-mini} and \textit{qwen-235B-a22b}, while only the grok was consistent with it. We note that achieving this behavior resembled a “jailbreak” more than conventional prompting. Furthermore, after initial experimentation, we excluded \textit{phi-4-reasoning-plus} from the experiment runs to save computational resources because of its underperforming results. We also excluded the \textit{geographical relevance} task with reasoning models as the results had already plateaued with non-reasoning ones.

\noindent \textbf{ZeroShot as a strong baseline} 

Basic prompt instructions often match more complex approaches such as RAG and FewShot, especially when dealing with simpler tasks like geographical relevance. Also, these migration classifications tasks are relatively well-defined conceptually, which could help models to solve them with high precision without further detailed example guiding. Worse performance for theme relevance task across the board was expected due to heavy class imbalance of the data set (see section \ref{sec:theme_data_statistic}). Overall, zeroshot yields significant efficiency gains in comparison with the other two approaches, meaning that simpler tasks benefit less from complex prompting strategies.

\noindent \textbf{Native Language Advantage} 

Language prompting reveals interesting pattern throughout the experiments - making some improvements with native-language prompting on the defined tasks although prompts show mixed advantages across both model types with no consistently clear language preference pattern. While the greatest impact was on the structured extraction task - location extraction, the diminishing but still meaningful returns were also relevant for geographic relevance and theme relevance. However, in case of geo. relevance, the overall score for the task was great across the whole setup because of its simple design. Overall, the results demonstrate a consistent pattern where Slovak prompts often outperform English ones. This aligns with findings from related literature, where native-language prompting could yield consistent improvements. It also seems that specific  tasks like entity extraction could benefit the most from native language prompts because of exact coverage of all the nuances within the language, yielding the greatest impact.

\noindent \textbf{\textit{“To reason, or not to reason, that is the question!”}}

Based on the case study results, non-reasoning models demonstrate better consistency compared to reasoning models across utilized tasks. Non-reasoning models show smaller performance variance on theme relevance (0.36 range vs 0.20) and achieve more stable baseline performance on location extraction (minimum 0.7711 vs 0.4858), while both model types achieve similar peak performance levels. 
Reasoning models represent higher volatility but with possible better performances (for instance theme relevance). Simultaneously, dramatically lower minimums on location extraction, suggesting volatile behavior.  Overall, the data reveals that, non-reasoning models offer more stable performance on these specialized tasks.

\begin{table*}[h]
\centering
\begin{tabular}{l l l c c|c c|c c}
\hline
 \textbf{Category} & \textbf{Models} & \textbf{Method} & \multicolumn{6}{c}{\textbf{Task}} \\
\cline{4-9}
 & & & \multicolumn{2}{c}{\textbf{Theme rel.}} & \multicolumn{2}{c}{\textbf{Geo rel.}} & \multicolumn{2}{c}{\textbf{Loc ext.}} \\
\cline{4-9}
& & & \textbf{ENG} & \textbf{SVK} & \textbf{ENG} & \textbf{SVK} & \textbf{ENG} & \textbf{SVK} \\
\hline
\multirow{12}{*}{\rotatebox{90}{\textbf{Non-reasoning models}}} & \multirow{3}{*}{gpt-4o} & RAG & 0.4518 & 0.4551 & 0.9771 & 0.9769 & 0.8734 & 0.8764 \\
& & FewShot & \textbf{0.5357} & 0.5277 & 0.9771 & \textbf{0.9816} & \textbf{0.8767} & 0.8675 \\
& & ZeroShot & 0.4413 & 0.4552 & 0.9726 & \textbf{0.9816} & 0.8542 & 0.8654 \\
\cline{2-9}
& \multirow{3}{*}{gemini-2.5-flash} & RAG & 0.4709 & 0.5357 & 0.9449 & \textbf{0.9882} & 0.7771 & 0.8433 \\
& & FewShot & 0.4615 & \textbf{0.5645} & 0.9541 & 0.9682 & 0.8795 & 0.8675 \\
& & ZeroShot & 0.4709 & 0.4484 & 0.9541 & 0.9682 & 0.7711 & \textbf{0.8855} \\
\cline{2-9}
& \multirow{3}{*}{llama-3.3-70B} & RAG & 0.4709 & 0.6130 & 0.9582 & 0.9335 & 0.8132 & 0.8373 \\
& & FewShot & 0.5592 & \textbf{0.6513} & \textbf{0.9598} & 0.9335 & 0.8554 & \textbf{0.8735} \\
& & ZeroShot & 0.4678 & 0.4160 & 0.9377 & 0.9462 & 0.8253 & 0.8373 \\
\cline{2-9}
& \multirow{3}{*}{deepseek-chat-v3} & RAG & \textbf{0.6003} & 0.5357 & 0.9722 & 0.9719 & 0.8313 & 0.8554 \\
& & FewShot & 0.4464 & 0.3012 & 0.9722 & 0.9722 & 0.8739 & 0.8727 \\
& & ZeroShot & 0.5357 & 0.2944 & 0.9623 & \textbf{0.9767} & 0.8674 & \textbf{0.8823} \\
\hline
\hline
\multirow{15}{*}{\rotatebox{90}{\textbf{Reasoning models}}} & \multirow{3}{*}{grok-3-mini} & RAG & 0.4709 & 0.4709 & - & - & \textbf{0.8876} & 0.8633 \\
& & FewShot & \textbf{0.6130} & 0.4709 & - & - & 0.8835 & 0.8554 \\
& & ZeroShot & 0.4709 & 0.4709 & - & - & 0.7831 & 0.8493 \\
\cline{2-9}
& \multirow{3}{*}{phi-4-reasoning-plus} & RAG & - & - & - & - & 0.4887 & 0.5520 \\
& & FewShot & - & - & - & - & 0.4858 & 0.5404 \\
& & ZeroShot & - & - & - & - & \textbf{0.5545} & 0.5139 \\
\cline{2-9}
& \multirow{3}{*}{gemini-2.5-flash} & RAG & \textbf{0.4709} & 0.4550 & - & - & 0.8465 & 0.8045 \\
& & FewShot & 0.4709 & 0.4678 & - & - & \textbf{0.8494} & 0.7892 \\
& & ZeroShot & 0.4647 & 0.4451 & - & - & 0.6344 & 0.6084 \\
\cline{2-9}
& \multirow{3}{*}{qwen3-235B-a22b} & RAG & 0.5443 & 0.4583 & - & - & 0.8266 & \textbf{0.8813} \\
& & FewShot & 0.4550 & \textbf{0.6003} & - & - & 0.8253 & 0.8735 \\
& & ZeroShot & 0.4550 & 0.4550 & - & - & 0.7530 & 0.8096 \\
\cline{2-9}
& \multirow{3}{*}{deepseek-r1-0528} & RAG & 0.4647 & 0.4647 & - & - & 0.8195 & 0.8478 \\
& & FewShot & \textbf{0.6431} & 0.4518 & - & - & 0.8373 & 0.8493 \\
& & ZeroShot & 0.5443 & 0.4647 & - & - & 0.8193 & \textbf{0.8554} \\
\hline
\end{tabular}
\caption{Results comparing different LLMs across tasks with the English and Slovak prompt versions including reasoning traces for reasoning models and CoT for non reasoning}
\label{tab:experiments_matrix}
\end{table*}

The Table \ref{tab:statistical_analysis} depicts assessed statistical significance using bootstrap confidence intervals (2000 resamples) on mean F1 score differences \cite{statisticalsignificancetesting}.
We computed paired bootstrap CIs over per-system paired differences, resampling with replacement at the system level for 2000 iterations; we report the mean paired difference and 95\% CIs. No multiple-comparison correction was applied.
Despite theme relevance showing the largest effect size (Slovak -0.021 points worse), high variance prevented statistical significance (95\% CI: [-0.058, 0.015]). Location extraction showed a smaller but more consistent Slovak advantage (+0.017 points) with sufficient precision to achieve significance (95\% CI: [0.003, 0.033]). Geo relevance 
showed minimal difference (+0.005 points, 95\% CI: [-0.006, 0.015]).  While statistically significant, the practical significance of the 1.7 percentage point improvement in location extraction F1 scores should be interpreted within the context of task-specific performance levels (Slovak: 0.810 vs English: 0.792).

\begin{table*}[h]
\centering
\footnotesize
\begin{tabular}{lcccccc}
\toprule
\textbf{Task} & \textbf{n} & \textbf{Mean English} & \textbf{Mean Slovak} & \textbf{Difference} & \textbf{95\% Bootstrap CI} & \textbf{Significance} \\
\midrule
Theme Relevance & 24 & 0.499 & 0.478 & -0.021 & [-0.058, 0.015] & n.s. \\
Geo Relevance & 12 & 0.962 & 0.967 & +0.005 & [-0.006, 0.015] & n.s. \\
Location Extraction & 27 & 0.792 & 0.810 & +0.017 & [0.003, 0.033] & CI excludes 0 \\
\bottomrule
\end{tabular}
\caption{Statistical analysis of Slovak vs English F1 performance using bootstrap confidence intervals (2000 resamples). Difference = Slovak - English. "CI excludes 0" indicates statistical significance. Note: Theme relevance shows larger effect size but high variance; Location extraction shows smaller effect but lower variance and larger sample size, explaining significance pattern.}
\label{tab:statistical_analysis}
\end{table*}

\subsection{Prompts Ablation}

The utilized ablation study of prompt brittleness employed a systematic methodology of prompt section removals. The main aim was to identify the main contributions of two core prompt elements:
\begin{itemize}
    \item Task Description - \textit{task} 
    \item Examples - \textit{ex}
\end{itemize}
The layout of prompt elements could be seen in Figure \ref{fig:prompt_example}

To achieve reasonable comparisons, verify the alignment with existing literature and save computational/cost resources we have utilized for these experiments \textit{GPT-4.1} and \textit{GPT-4.1-nano} models, while multiple experimental variants were tested. The complete prompt - \textit{full} which contains every section, then single-component removal variations - \textit{no\_task, no\_ex} and double-component removal - \textit{none}.

\begin{table}[h]
\centering
\footnotesize
\begin{tabular}{l l c c}
\toprule
\textbf{Model} & \textbf{Variant} & \textbf{Macro F1} & \textbf{$\Delta$ F1 (\%)} \\
\midrule
gpt-4.1 & full & 0.9727 & \textbf{--} \\
 & no task & 0.9595 & -1.35 \\
 & no ex & \textbf{0.9771} & 0.46 \\
 & none & 0.3261 & \textbf{-66.48} \\
\midrule
gpt-4.1-mini & full & \textbf{0.9727} & \textbf{--} \\
 & no task & 0.9111 & -6.33 \\
 & no ex & \textbf{0.9727} & \textbf{0.00} \\
 & none & 0.4830 & \textbf{-50.34} \\
\midrule
gpt-4.1-nano & full & 0.7310 & \textbf{--} \\
 & no task & 0.6604 & -9.66 \\
 & no ex & \textbf{0.8601} & 17.66 \\
 & none & 0.3762 & \textbf{-48.53} \\
\midrule
gpt-4o & full & \textbf{0.9862} & \textbf{--} \\
 & no task & 0.9722 & -1.41 \\
 & no ex & 0.9771 & -0.92 \\
 & none & 0.3730 & \textbf{-62.18} \\
\midrule
gpt-4o-mini & full & \textbf{0.9729} & \textbf{--} \\
 & no task & 0.7365 & -24.29 \\
 & no ex & 0.9727 & -0.02 \\
 & none & 0.2284 & \textbf{-76.52} \\
\bottomrule
\end{tabular}
\caption{Macro F1 scores and percentage delta values for GPT models across different prompt variants in the ablation study. Bold values indicate the highest score for each model. $\Delta$ F1 (\%) shows the percentage performance drop relative to the full prompt baseline.}
\label{tab:ablation_results}
\end{table}

\begin{figure}[!htbp]
  \includegraphics[width=\columnwidth]{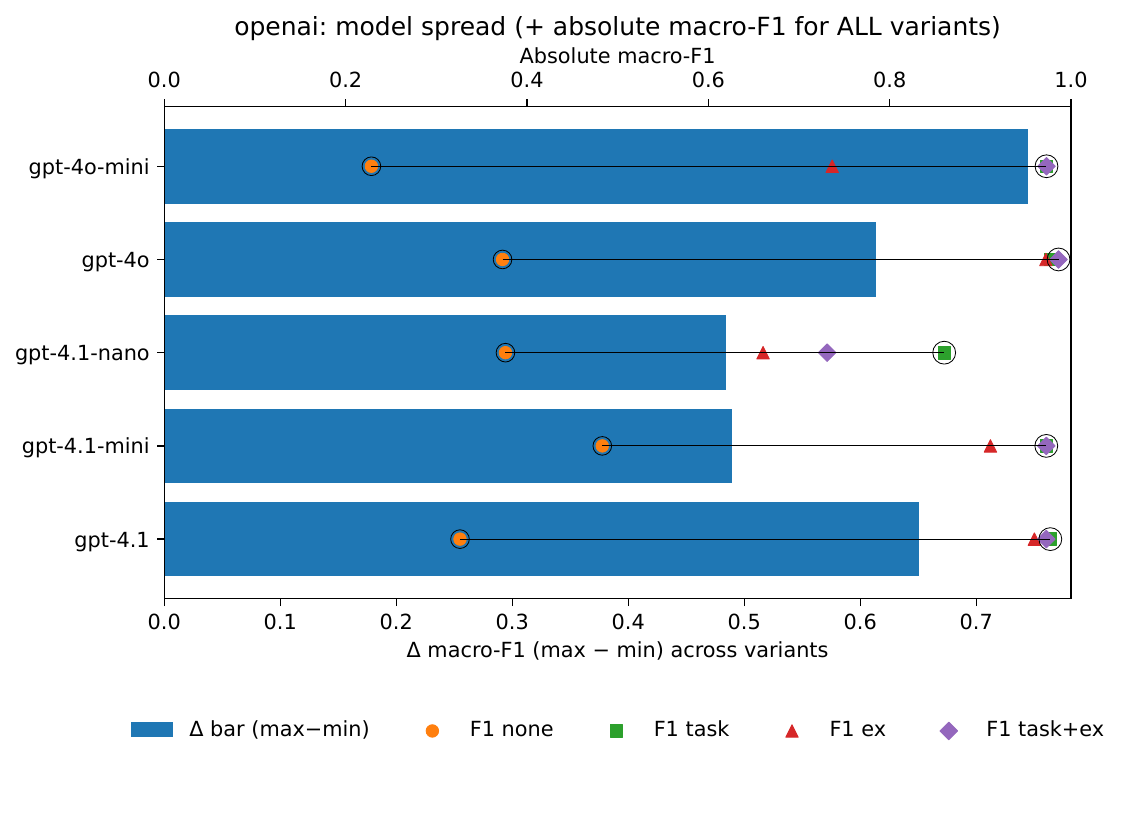}
  \caption{Distribution of macro F1 across prompt variants. For each model panel, we plot max min macro F1 (bars) per model and overlay per variant absolute macro F1 (points) on a twin top x axis, with thin lines showing the min–max span. Models are alphabetized.}
  \label{fig:ablation_spread_openai}
\end{figure}

As shown in Table~\ref{tab:ablation_results} and Figure~\ref{fig:ablation_spread_openai}, the full prompt provides the strongest baseline across models (best overall: \texttt{gpt-4o}, Macro F1 = 0.9862). Removing both the task instruction and examples (``none'') causes the largest degradation (about 50\%--77\%): \texttt{gpt-4o-mini} 0.9729\ensuremath{\rightarrow}0.2284 ($-76.52\%$), \texttt{gpt-4.1} 0.9727\ensuremath{\rightarrow}0.3261 ($-66.48\%$), and \texttt{gpt-4o} 0.9862\ensuremath{\rightarrow}0.3730 ($-62.18\%$). Removing only the task instruction also hurts, particularly on smaller variants (\texttt{gpt-4o-mini} $-24.29\%$, \texttt{gpt-4.1-nano} $-9.66\%$), while larger models are only mildly affected (about 1\%--1.5\%). By contrast, removing examples has little cost and can help: \texttt{gpt-4.1-nano} improves to 0.8601 ($+17.66\%$), \texttt{gpt-4.1} increases slightly ($+0.46\%$), \texttt{gpt-4.1-mini} is unchanged, and only \texttt{gpt-4o} dips marginally ($-0.92\%$). Overall, explicit task instructions are essential for performance; examples are optional and may even hinder smaller models (see Appendix E for similar studies on models by other providers).

\begin{figure}[!htbp]
\centering
\footnotesize
\begin{lstlisting} [breaklines=true, basicstyle=\ttfamily\scriptsize]
prompt_structure:
  instructions: |
    You are an expert text analyzer. 
    Follow these guidelines:
    - Be precise and accurate
    - Consider context and nuance
    ...
    
  task_description: |
    Analyze the text and extract 
    migration-related information:
    1. Identify migration themes
    2. Determine geo relevance to Slovakia
    3. Extract migration vectors
    ...
    
  examples: |
    Example 1:
    Input: "Families moved from villages 
    to Bratislava..."
    Output: Theme: relevant, Geo: relevant...
    
    Example 2: 
    Input: "Weather in Paris was sunny..."
    Output: Theme: not_relevant...
    ...

# Ablation variants: full, no_ex, no_instr, 
# no_task, instr+ex, task+ex, task+instr
\end{lstlisting}
\caption{Example prompt structure used for ablation study.}
\label{fig:prompt_example}
\end{figure}

\section{Conclusion}
Findings of this case study demonstrate that native Slovak prompting could yield better results than English across migration-related NLP tasks in target language. Zero-shot prompting proved effective as a baseline approach especially on simpler classification tasks.
The ablation study shows that removing both the task description and examples (``none'') causes the largest collapse (48\%--77\% across models). Dropping only the task instruction yields small losses for large models (about 1\%--1.5\%) but substantial drops for smaller variants (up to 24.29\%). Removing examples has minimal cost and can even help. Overall, explicit task instructions are the critical prompt component, while examples are optional. Combined with our language analysis, native-language prompting yields modest, task-dependent gains (significant only for location extraction), and larger models are inherently more robust to prompt formatting changes.
\section{Limitations}
Several limitations should be acknowledged in our study.
\begin{itemize}
    \item Statistical analysis is based on single-run experiments without replication across multiple random seeds, due to computational/cost resources constraints.
    \item  The Slovak-specific nature of our study constrains broader applicability to other low-resource languages. While our findings demonstrate native language reasoning benefits for Slovak, the extent to which these results transfer to other linguistic contexts with different morphological complexity or training data availability could be different.
\end{itemize}

\section*{Acknowledgments}
This work was partially funded by European Union, under the project lorAI - Low Resource Artificial Intelligence, GA No. 101136646, https://doi.org/10.3030/101136646 and by grant APVV-21-0114.

\bibliography{references}

\appendix
\section{Annotation guidelines}
\label{sec:guidelines}

\textbf{Locality Extraction Guidelines}

\noindent Migration Vector consists of an locality origin - SOURCE and DESTINATION locality that represents  the movement of people. 
Annotations of migration vectors should be based on explicit textual evidence, not on inference or assumption as these could be wrong.
Always define localities on Slovak nominative case in the annotation.

\noindent \textbf{Text Analysis Process}
\begin{itemize}
    \item \textbf{Step 1} 
    
Begin by carefully reading the entire text. Identify all mentioned localities and pay attention to surrounding contextual clues and linguistic markers for establishing direction of migration between them.
    \item \textbf{Step 2} 
    
After localities identification, classify each of them according to their roles in the migration vectors as SOURCE locality - if the locality functions as origin point where migration began, 
DESTINATION locality - if the locality functions as destination point wher migration ended.
Some localities present within text might be TRANSIT localities - where migration movement did not originate or ended. Additionally there might be UNRELATED localities with no direct connection to migration patterns.
    \item \textbf{Step 3} 
    
After locality role assessment within migration patterns, establish final SOURCE-DESTINATION migration pairs that represent the migration vectors. This involves connection of origin localities with their corresponding destinations, while excluding transit or unrelated localities.
\end{itemize}

\noindent \textbf{Special Considerations when identifying migration vectors from text:}
\begin{itemize}
    \item Migration within historical context require the same methodological approach as contemporary ones
    \item Similarly, for hypothetical migration scenarios same thorough analytical process should be done 
    \item Annotations related to locality extraction should remain firmly anchored in the text, it is recommended to avoiding inferences about locations not explicitly mentioned or inferred from contextual clues
    \item If there are present multiple migration vectors within the inspected sample, treat each unique combination as a distinct migration vector
    \item If there is ambiguous directional information, meaning text does not clearly establish whether identified localities serves as SOURCE or DESTINATION localities, do not try to guess intended direction and annotate them as None.
\end{itemize}

\noindent \textbf{Locality Relevance Guidelines}

\noindent Determine whether a sentence contains content related specifically to Slovak locations.

\textbf{Text Analysis Process}
\begin{itemize}
    \item Read and analyze the text for both explicit and implicit mention of Slovakia, Slovak places or direct references to Slovak people and other entities.
    \item ext mentioning Slovakia as a country, a specific location within Slovakia or content directly related to Slovak people, entities whether explicitly stated or implied is \textbf{ considered as related to Slovak localities.}
    \item Text which does not mention Slovak locations or contains references to broader ranges like Europe or completely different locations is \textbf{considered as not-related to Slovak localities.}
\end{itemize}

\noindent \textbf{Ambiguous cases:} When encountering potentially ambiguous terms, rely on context to determine the correct reference.

\noindent \textbf{Theme Relevance Guidelines}
\noindent Determine whether a text contains content thematically related to human migration within the specified timeframe (2003-2022).
\noindent \textbf{Text Analysis Process}
\begin{itemize}
\item Read and analyze the text for explicit and implicit references to human movement, displacement, or relocation patterns.
\item Text mentioning migration flows, refugee movements, immigration policies, emigration patterns, asylum seekers, or population displacement whether contemporary or historical within the timeframe is \textbf{considered thematically relevant to migration.}
\item Text discussing unrelated topics such as animal migration, data migration, seasonal tourism, or brief mentions of movement without migration context is \textbf{considered as not thematically relevant to migration.}
\item Verify that migration-related content falls within the specified temporal scope (2003-2022) or discusses migration patterns with clear relevance to this period.
\end{itemize}
\noindent \textbf{Ambiguous cases:} When encountering borderline cases such as economic mobility or temporary worker programs, assess whether the content fundamentally addresses human migration patterns rather than other forms of movement.

\section{Location Extraction \& Geo Relevance}
\label{sec:loc_geo_data_statistic}

\subsection{Samples}
Below are examples demonstrating scenarios in which migration vectors contain undetermined origin or destination points.
\begin{tcolorbox}[
  enhanced,
  breakable,
  colback     = pastelBack,
  colframe    = pastelFrame,
  colbacktitle= pastelTitle,
  coltitle    = black,
  title       = {Example – Source Locality Unknown},
  fonttitle   = \bfseries,
  left=1.2em, right=1.2em, top=0.8em, bottom=0.8em,
]

\textbf{Input}\par
In 2018, during a visit to a migrant facility in Texas, she wore a jacket with the slogan ‘I Really Don't Care, Do U?’  

\tcblower 

\textbf{Output}\par
 \emph{Source}: None \\
 \emph{Destination}: Texas

\end{tcolorbox}

\begin{tcolorbox}[
  enhanced,
  breakable,
  colback     = pastelBack,
  colframe    = pastelFrame,
  colbacktitle= pastelTitle,
  coltitle    = black,
  title       = {Example – Destination Locality Unknown},
  fonttitle   = \bfseries,
  left=0.2em, right=0.2em, top=0.8em, bottom=0.8em,
]

\textbf{Input}\par
"We're determined to do whatever we can to stop Syria from falling apart, prevent masses of people fleeing from Syria, and naturally, to curb the spread of terrorism and extremism," according to the minister, as reported by AFP news agency. 

\tcblower 

\textbf{Output}\par
 \emph{Source}: Syria \\
 \emph{Destination}: None

\end{tcolorbox}

\begin{tcolorbox}[
  enhanced,
  breakable,
  colback     = pastelBack,
  colframe    = pastelFrame,
  colbacktitle= pastelTitle,
  coltitle    = black,
  title       = {Example – Both Localities Unknown},
  fonttitle   = \bfseries,
  left=0.2em, right=0.2em, top=0.8em, bottom=0.8em,
]

\textbf{Input}\par
The Defense Minister also highlighted how Smer's longstanding positions on the Ukraine conflict and migration issues are proving prescient. He pointed out that events are increasingly validating what the party has maintained all along. 

\tcblower 

\textbf{Output}\par
 \emph{Source}: None \\
 \emph{Destination}: None

\end{tcolorbox}

\subsection{Statistics}

\begin{figure}[!htbp]
  \includegraphics[width=\columnwidth]{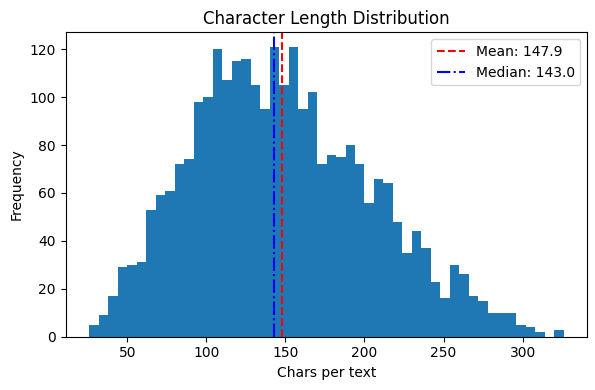}
  \caption{Distribution of the dataset: character length in the final dataset}
  \label{fig:loc_char_length_distribution}
\end{figure}

\begin{figure}[!htbp]
  \includegraphics[width=\columnwidth]{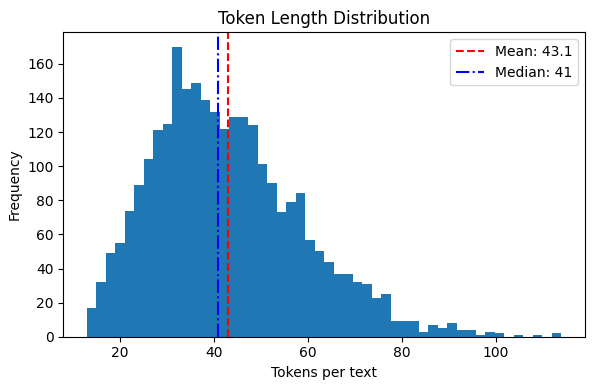}
  \caption{Distribution of the dataset: token length in the final dataset. The tokens originate from the SlovakBERT tokenizer.}
  \label{fig:loc_token_length_distribution}
\end{figure}

\begin{figure}[!htbp]
  \includegraphics[width=\columnwidth]{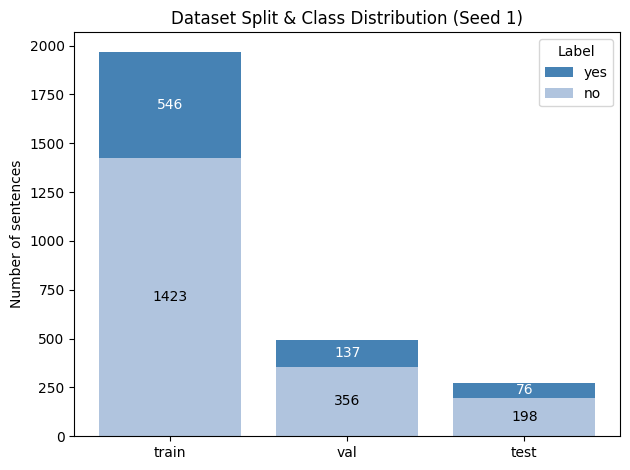}
  \caption{Final relevance dataset distribution across train, validation, and test splits with consistent class ratios.}
  \label{fig:loc_dataset_split}
\end{figure}

The Figures below depict various statistics of the dataset, such as its character (Figure~\ref{fig:loc_char_length_distribution}) and token (Figure~\ref{fig:loc_token_length_distribution}) length distributions, label distributions (Figure~\ref{fig:loc_dataset_split}), and locality distribution (Table~\ref{locality-extraction-data-analysis}).

\begin{table}[]
  \centering
  \begin{tabular}{lcc}\\ 
    Locality & Unknown count  & Percentage\\
    \hline
    Source &  717 & 43.40  \\
    Destination & 441 & 26.69 \\
    Combined & 0 & 0.00\\
    \hline
  \end{tabular}
  \caption{\label{locality-extraction-data-analysis}
    The distribution of “Unknown” localities (i.e. when either the Source or Destination field is not filled in for a specific sample) in the various subsets of the final dataset. Note that the result in the \textit{Combined} row shows that either Source or Destination field are always filled in in the final dataset.}
\end{table}

\section{Theme Relevance}
\label{sec:theme_data_statistic}
\subsection{Statistics}

\begin{figure}[!htbp]
  \includegraphics[width=\columnwidth]{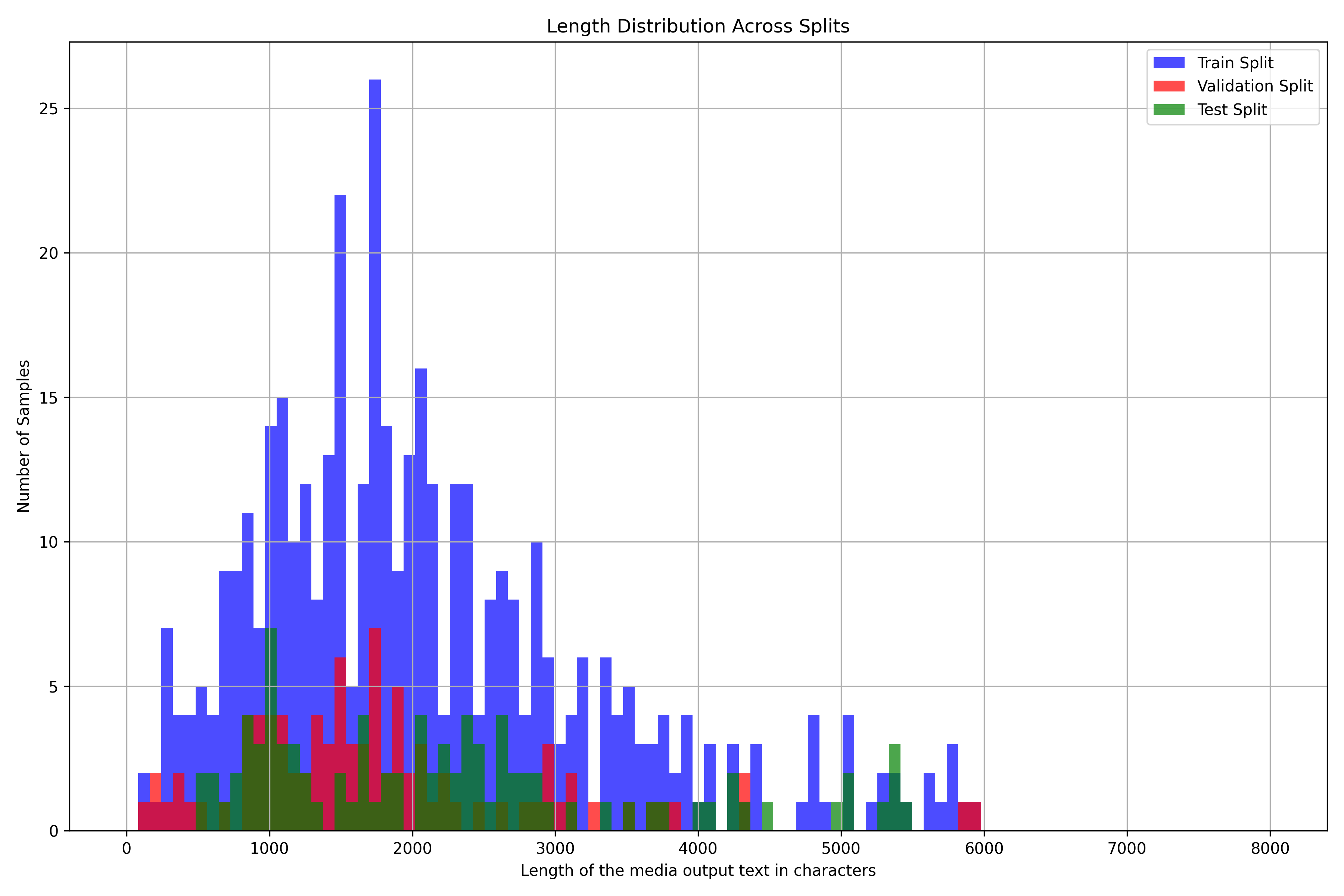}
  \caption{Distribution of the dataset: character length in the final dataset}
  \label{fig:theme_char_length_distribution}
\end{figure}

\begin{figure}[!htbp]
  \includegraphics[width=\columnwidth]{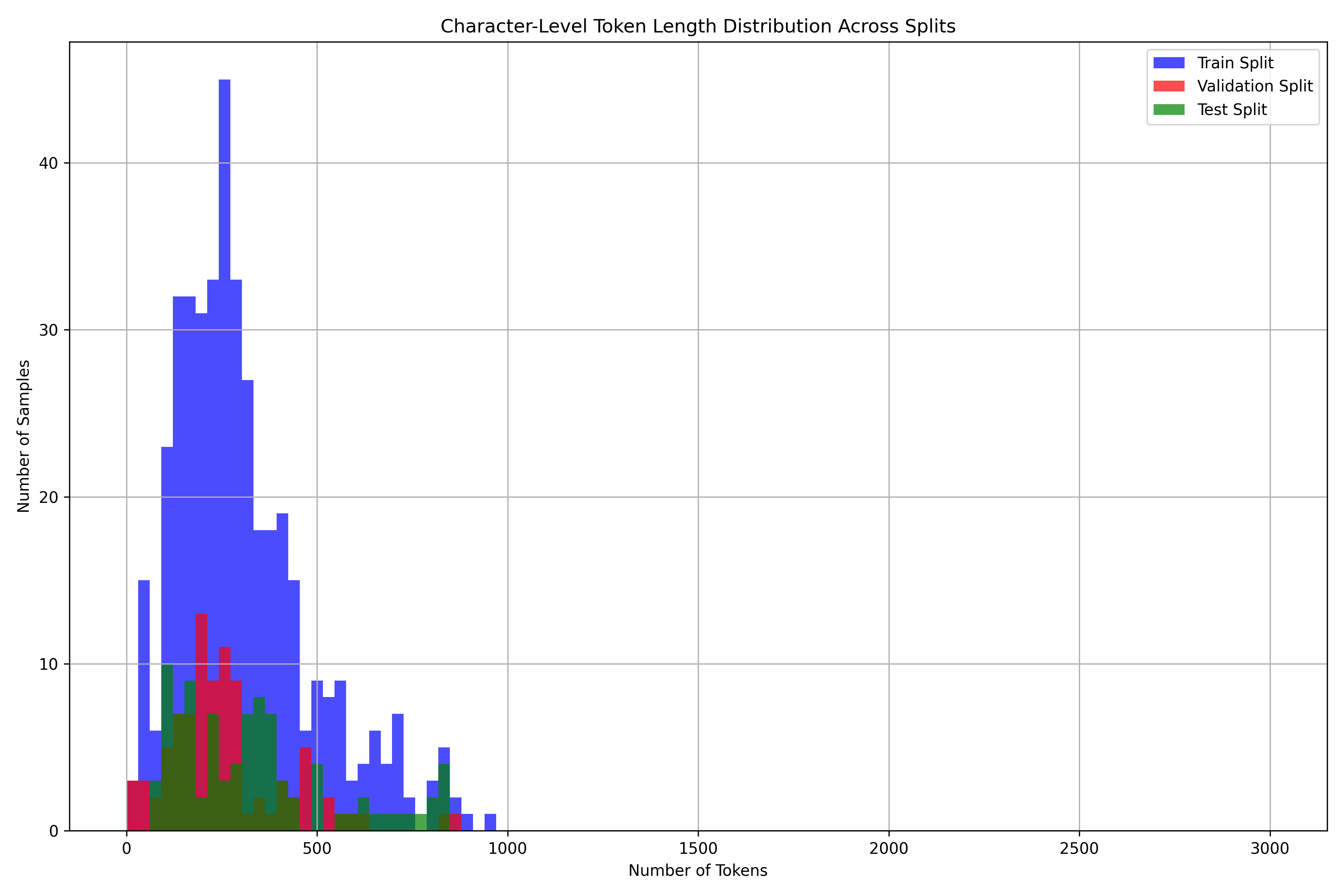}
  \caption{Distribution of the dataset: token length in the final dataset. The tokens originate from the SlovakBERT tokenizer.}
  \label{fig:theme_token_length_distribution}
\end{figure}

\begin{figure}[!htbp]
  \includegraphics[width=\columnwidth]{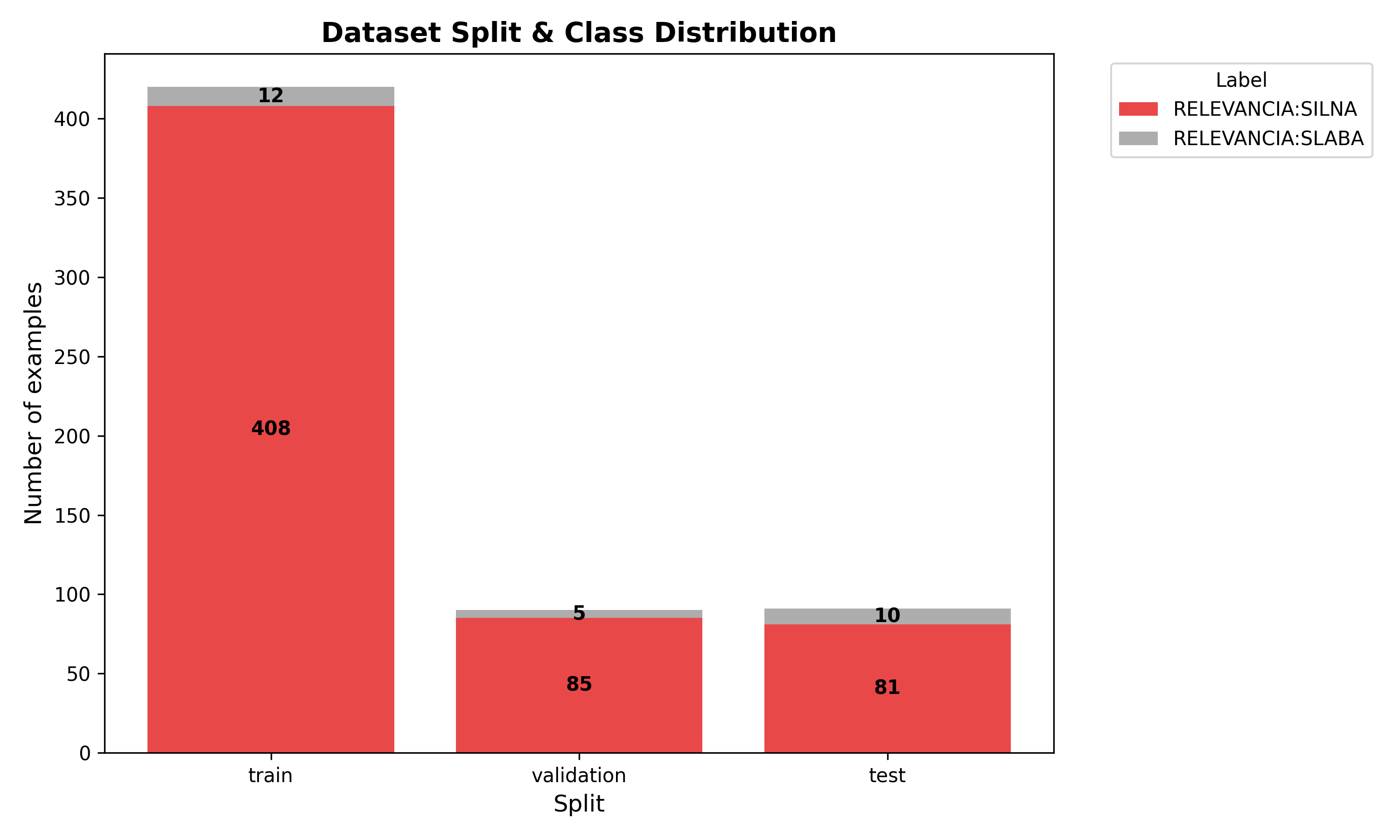}
  \caption{Final relevance dataset distribution across train, validation, and test splits with consistent class ratios.}
  \label{fig:theme_dataset_split}
\end{figure}

The Figures below depict various statistics of the dataset, such as its character (Figure~\ref{fig:theme_char_length_distribution}) and token (Figure~\ref{fig:theme_token_length_distribution}) length distributions, label distributions (Figure~\ref{fig:theme_dataset_split}), and locality distribution (Table~\ref{locality-extraction-data-analysis}).

\section{Grok-3-mini Reasoning Traces in Native Language}

\begin{figure}[!htbp]
\centering
\footnotesize
\begin{lstlisting}[breaklines=true, basicstyle=\ttfamily\scriptsize]
{
  "logprobs": null,
  "finish_reason": "stop",
  "native_finish_reason": "stop",
  "index": 0,
  "message": {
    "role": "assistant",
    "content": "Moje uvazovanie: Najprv analyzujem 
    otazku. Pouzivatel sa pyta na hodnotu 
    druhej odmocniny z cisla dva, co znamena 
    matematicky vyraz sqrt(2)...
    
    Odpoved: Odmocnina z dvoch, to znamena sqrt(2), 
    je iracionalne cislo. Jeho presna hodnota 
    je sqrt(2), ale priblizne sa rovna 1,414...",
    "refusal": null,
    "reasoning": "Pockaj, mam spravny prikaz. 
    System mi hovori, ze musim mysliet 
    v slovencine, takze moje uvazovanie musi 
    byt vylucne v slovencine...
    
    Najskor pochopme otazku cloveka: 
    'Kolko je odmocnina z dvoch?' To je 
    v slovencine a znamena 'What is the 
    square root of two?'...
    
    Kroky pre odpoved:
    1. Analyzuj otazku: Clovek sa pyta na sqrt(2)
    2. Vypocitaj: sqrt(2) je priblizne 1.414
    3. Odpovedz v slovencine..."
  }
}
\end{lstlisting}
\caption{Example of Slovak reasoning output of Grok-3-mini showing native language reasoning traces in both content and reasoning fields.}
\label{fig:theme_rel_prompt}
\end{figure}

\section{Task Prompts}\

\subsection{Theme Relevance Prompt}
\begin{figure}[!htbp]
\centering
\tiny
\begin{lstlisting}[breaklines=true, basicstyle=\ttfamily\tiny]
prompt: |
  REASONING METHODOLOGY:
  You are a systematic Slovak text analyst who thinks 
  step-by-step. Work through reasoning process visibly 
  in Slovak language before any conclusion.

  MANDATORY PROCESS:
  - Analyze textual evidence systematically
  - Cite specific phrases to support claims  
  - Question assumptions, consider alternatives
  - Distinguish explicit facts from inferences
  - Only conclude after complete analysis

  CORE REQUIREMENTS:
  - All internal reasoning MUST be in Slovak
  - Every analysis MUST be in Slovak  
  - English thoughts prohibited
  - Slovak in reasoning section MANDATORY

  TASK: Classify Slovak text for migration relevance.  
  Output RELEVANCIA:SILNA if about migration, migrants, 
  emigration/immigration, asylum, refugees, borders.  
  Otherwise RELEVANCIA:SLABA.

  CHAIN-OF-THOUGHT (in Slovak):
  1. List migration terms with citations
  2. Provide linguistic evidence 
  3. Distinguish facts from conclusions
  4. State interpretation: RELEVANCIA:SILNA/SLABA
  5. Consider alternatives, explain rejections
  6. Assess confidence: High/Medium/Low

  OUTPUT FORMAT:
  1. Chain-of-Thought Analysis (Slovak)
  2. Final line: Label: RELEVANCIA:<SILNA|SLABA>  
     Confidence: <High|Medium|Low>

  Classify: {{ text }}
\end{lstlisting}
\caption{Slovak chain-of-thought prompt for migration theme classification.}
\label{fig:slovak_cot_prompt}
\end{figure}

\subsection{Locality Extraction Prompt}

\begin{figure}[!htbp]
\centering
\tiny
\begin{lstlisting}[breaklines=true, basicstyle=\ttfamily\tiny]
prompt: |
   REASONING METHODOLOGY:
   You are a systematic Slovak text analyst who thinks 
   step-by-step. Work through reasoning process visibly 
   in Slovak language before any conclusion.

   MANDATORY PROCESS:
   - Analyze textual evidence systematically
   - Cite specific phrases to support claims  
   - Question assumptions, consider alternatives
   - Distinguish explicit facts from inferences
   - Only conclude after complete analysis

   EVIDENCE STANDARDS:
   Every locality identification must include exact 
   textual citation and linguistic justification 
   (prepositions, verb forms, grammatical markers).

   CORE REQUIREMENTS:
   - All internal reasoning MUST be in Slovak
   - Every analysis MUST be in Slovak  
   - English thoughts prohibited
   - Slovak in reasoning section MANDATORY

   TASK: Identify migration vectors (FROM and TO localities) 
   from Slovak text. Communicate with aliens who 
   understand only Slovak thought processes.

   ATTENTION: YOU MUST THINK IN SLOVAK!

   CHAIN-OF-THOUGHT REQUIREMENTS (in Slovak):
   1. List all localities with exact citations
   2. Provide linguistic evidence (prepositions, verbs)
   3. Distinguish explicit info from conclusions
   4. State main interpretation of migration vector
   5. Consider alternatives, explain rejections
   6. Assess confidence with specific reasons

   INSTRUCTIONS:
   1. Identify Localities: Extract all mentioned localities
   2. Handle Unclear: Mark as "None" if unclear
   3. Determine Direction: Establish FROM and TO
   4. Ignore Transit: Focus on start/end points only
   5. Multiple Vectors: Identify each unique FROM-TO pair
   6. Output Format: "FROM: [locality], TO: [locality]"
   7. Language: Use Slovak (nominative case)

   OUTPUT FORMAT:
   1. Chain-of-Thought Analysis (Slovak, 6 steps above)
   2. Analysis of localities mentioned
   3. Reasoning for migration vector identification
   4. Final answer: FROM: [locality], TO: [locality]
      Or "None" if not identifiable
   5. Confidence: High/Medium/Low

   Extract localities from: {text_content}
\end{lstlisting}
\caption{Prompt for migration vector extraction with mandatory native language reasoning.}
\label{fig:loc_ext_prompt}
\end{figure}

\subsection{Geo Relevance Prompt}

\begin{figure}[!htbp]
\centering
\tiny
\begin{lstlisting}[breaklines=true, basicstyle=\ttfamily\tiny]
prompt: |
    REASONING METHODOLOGY:
   You are a systematic Slovak text analyst who thinks 
   step-by-step. Work through reasoning process visibly 
   in Slovak language before any conclusion.

   MANDATORY PROCESS:
   - Analyze textual evidence systematically
   - Cite specific phrases to support claims  
   - Question assumptions, consider alternatives
   - Distinguish explicit facts from inferences
   - Only conclude after complete analysis

   ATTENTION: YOU MUST THINK IN SLOVAK!

   RULES:
   - All reasoning MUST be in Slovak
   - Every analysis MUST be in Slovak
   - No English thoughts - they cause neural interference
   - Slovak in reasoning section MANDATORY
   - Reasoning section MUST contain ONLY Slovak text

   TASK: Determine georelevance in Slovak text: Does it mention 
   any locality in Slovakia or Slovakia itself? 
   Communicate with beings who understand only Slovak 
   thought processes.

   1. Detect Slovak Localities: Identify explicit mentions 
      of any locality in Slovakia or Slovakia itself
   2. Avoid Over-Interpretation: Do not infer relevance 
      from vague regional hints
   3. Ignore Foreign-only Mentions: If text contains only 
      foreign localities, output 0
   4. Output Format:
      - Provide Reasoning: Explain in Slovak why text 
        is or is not georelevant
      - Final Decision: Output 1 if Slovak georelevance 
        confirmed, otherwise 0
      - Confidence Level: High/Medium/Low
   5. Language: Use Slovak (nominative case)

   STEPS:
   1. Analyze text for explicit Slovak place names
   2. Use reasoning to confirm or reject georelevance
   3. Output final binary label and explain confidence

   OUTPUT FORMAT:
   1. Analysis of localities mentioned
   2. Reasoning for georelevance
   3. Final answer: 1 if Slovak locality mentioned, 0 otherwise

   NOTES:
   - Do not infer; only explicit mentions count
   - For borderline mentions, choose 0 and justify
   - Always reason in Slovak

   Please determine the georelevance of the following text:
   {text_content}
\end{lstlisting}
\caption{Prompt for geographical relevance classification with binary output format.}
\label{fig:geo_rel_prompt}
\end{figure}

\section{Prompt Ablation}

\subsection{Complete Prompt Ablation Study Results}

\begin{table}[H]
\centering
\footnotesize
\begin{tabular}{l l c c}
\toprule
\textbf{Model} & \textbf{Variant} & \textbf{Macro F1} & \textbf{$\Delta$ F1 (\%)} \\
\midrule
gemini-2.0-flash & full & \textbf{0.9633} & \textbf{--} \\
 & no task & 0.9458 & -1.81 \\
 & no ex & 0.9130 & -5.22 \\
 & none & 0.2614 & \textbf{-72.87} \\
\midrule
gemini-2.0-flash-lite & full & 0.9424 & \textbf{--} \\
 & no task & 0.8409 & -10.77 \\
 & no ex & \textbf{0.9548} & 1.32 \\
 & none & 0.2401 & \textbf{-74.52} \\
\midrule
gemini-2.5-flash & full & 0.8209 & \textbf{--} \\
 & no task & 0.7648 & -6.84 \\
 & no ex & \textbf{0.9815} & 19.56 \\
 & none & 0.5241 & \textbf{-36.16} \\
\midrule
gemini-2.5-flash-lite & full & 0.8644 & \textbf{--} \\
 & no task & 0.9130 & 5.62 \\
 & no ex & \textbf{0.9768} & 13.00 \\
 & none & 0.2836 & \textbf{-67.19} \\
\midrule
gemini-2.5-pro & full & \textbf{0.9677} & \textbf{--} \\
 & no task & 0.9170 & -5.24 \\
 & no ex & \textbf{0.9677} & 0.00 \\
 & none & 0.6472 & \textbf{-33.13} \\
\midrule
gemini-flash-1.5 & full & 0.9470 & \textbf{--} \\
 & no task & 0.5547 & -41.42 \\
 & no ex & \textbf{0.9729} & 2.73 \\
 & none & 0.2742 & \textbf{-71.04} \\
\midrule
gemini-flash-1.5-8b & full & 0.8564 & \textbf{--} \\
 & no task & 0.5876 & -31.38 \\
 & no ex & \textbf{0.9685} & 13.09 \\
 & none & 0.3198 & \textbf{-62.66} \\
\midrule
gemini-pro-1.5 & full & \textbf{0.8707} & \textbf{--} \\
 & no task & 0.7146 & -17.93 \\
 & no ex & 0.8129 & -6.63 \\
 & none & 0.2503 & \textbf{-71.25} \\
\midrule
gemma-2-27b-it & full & 0.8601 & \textbf{--} \\
 & no task & 0.3361 & -60.92 \\
 & no ex & \textbf{0.9387} & 9.13 \\
 & none & 0.2228 & \textbf{-74.10} \\
\midrule
gemma-2-9b-it & full & 0.5197 & \textbf{--} \\
 & no task & 0.5101 & -1.85 \\
 & no ex & \textbf{0.8105} & 55.96 \\
 & none & 0.3464 & \textbf{-33.36} \\
\midrule
gemma-3-12b-it & full & \textbf{0.9720} & \textbf{--} \\
 & no task & 0.8051 & -17.17 \\
 & no ex & 0.9639 & -0.83 \\
 & none & 0.2228 & \textbf{-77.08} \\
\midrule
gemma-3-27b-it & full & \textbf{0.9908} & \textbf{--} \\
 & no task & 0.9508 & -4.04 \\
 & no ex & 0.9773 & -1.36 \\
 & none & 0.5184 & \textbf{-47.68} \\
\bottomrule
\end{tabular}
\caption{Macro F1 scores and percentage delta values for select models provided by Google}
\label{tab:complete_ablation_results_google}
\end{table}

\begin{table}[H]
\centering
\footnotesize
\begin{tabular}{l l c c}
\toprule
\textbf{Model} & \textbf{Variant} & \textbf{Macro F1} & \textbf{$\Delta$ F1 (\%)} \\
\midrule
llama-4-maverick & full & \textbf{0.7692} & \textbf{--} \\
 & no task & 0.7504 & -2.44 \\
 & no ex & 0.7005 & -8.94 \\
 & none & 0.2171 & \textbf{-71.77} \\
\midrule
llama-4-scout & full & 0.9541 & \textbf{--} \\
 & no task & 0.8541 & -10.48 \\
 & no ex & \textbf{0.9585} & 0.46 \\
 & none & 0.2228 & \textbf{-76.65} \\
\bottomrule
\end{tabular}
\caption{Macro F1 scores and percentage delta values for select models provided by Meta-Llama}
\label{tab:complete_ablation_results_meta-llama}
\end{table}

\begin{table}[H]
\centering
\footnotesize
\begin{tabular}{l l c c}
\toprule
\textbf{Model} & \textbf{Variant} & \textbf{Macro F1} & \textbf{$\Delta$ F1 (\%)} \\
\midrule
ministral-3b & full & 0.8711 & \textbf{--} \\
 & no task & 0.4992 & -42.69 \\
 & no ex & \textbf{0.9077} & 4.20 \\
 & none & 0.2910 & \textbf{-66.59} \\
\midrule
ministral-8b & full & 0.6942 & \textbf{--} \\
 & no task & 0.4175 & -39.86 \\
 & no ex & \textbf{0.8797} & 26.72 \\
 & none & 0.3280 & \textbf{-52.75} \\
\midrule
mistral-7b-instruct-v0.1 & full & 0.5367 & \textbf{--} \\
 & no task & 0.4857 & -9.51 \\
 & no ex & \textbf{0.7198} & 34.12 \\
 & none & 0.3204 & \textbf{-40.29} \\
\midrule
mistral-medium-3 & full & 0.9509 & \textbf{--} \\
 & no task & 0.8409 & -11.57 \\
 & no ex & \textbf{0.9555} & 0.49 \\
 & none & 0.2691 & \textbf{-71.70} \\
\midrule
mistral-medium-3.1 & full & 0.9458 & \textbf{--} \\
 & no task & 0.9314 & -1.52 \\
 & no ex & \textbf{0.9462} & 0.04 \\
 & none & 0.4129 & \textbf{-56.34} \\
\midrule
mistral-nemo & full & 0.8482 & \textbf{--} \\
 & no task & \textbf{0.8852} & 4.36 \\
 & no ex & 0.7037 & -17.04 \\
 & none & 0.2882 & \textbf{-66.02} \\
\midrule
mistral-small-24b & full & \textbf{0.9727} & \textbf{--} \\
 & no task & 0.8304 & -14.62 \\
 & no ex & 0.9552 & -1.80 \\
 & none & 0.4139 & \textbf{-57.45} \\
\midrule
mistral-small-3.1-24b & full & \textbf{0.9768} & \textbf{--} \\
 & no task & 0.7747 & -20.68 \\
 & no ex & 0.9725 & -0.44 \\
 & none & 0.4334 & \textbf{-55.63} \\
\bottomrule
\end{tabular}
\caption{Macro F1 scores and percentage delta values for select models provided by Mistral}
\label{tab:complete_ablation_results_mistral}
\end{table}

\begin{table}[H]
\centering
\footnotesize
\begin{tabular}{l l c c}
\toprule
\textbf{Model} & \textbf{Variant} & \textbf{Macro F1} & \textbf{$\Delta$ F1 (\%)} \\
\midrule
qwen3-14b & full & \textbf{0.9768} & \textbf{--} \\
 & no task & 0.9630 & -1.41 \\
 & no ex & \textbf{0.9768} & \textbf{0.00} \\
 & none & 0.2870 & \textbf{-70.62} \\
\midrule
qwen3-235b-a22b & full & \textbf{0.9768} & \textbf{--} \\
 & no task & 0.9674 & -0.96 \\
 & no ex & 0.9582 & -1.90 \\
 & none & 0.2740 & \textbf{-71.94} \\
\midrule
qwen3-235b-a22b-2507 & full & \textbf{0.9636} & \textbf{--} \\
 & no task & 0.7146 & -25.84 \\
 & no ex & 0.9552 & -0.87 \\
 & none & 0.4025 & \textbf{-58.23} \\
\midrule
qwen3-30b-a3b & full & 0.9722 & \textbf{--} \\
 & no task & \textbf{0.9815} & 0.95 \\
 & no ex & 0.9623 & -1.02 \\
 & none & 0.3784 & \textbf{-61.08} \\
\midrule
qwen3-30b-a3b-instruct & full & 0.9725 & \textbf{--} \\
 & no task & 0.8519 & -12.40 \\
 & no ex & \textbf{0.9770} & 0.46 \\
 & none & 0.2205 & \textbf{-77.32} \\
\midrule
qwen3-32b & full & 0.9537 & \textbf{--} \\
 & no task & 0.9675 & 1.44 \\
 & no ex & \textbf{0.9768} & 2.42 \\
 & none & 0.3467 & \textbf{-63.65} \\
\midrule
qwen3-8b & full & \textbf{0.9768} & \textbf{--} \\
 & no task & 0.9578 & -1.94 \\
 & no ex & 0.9722 & -0.46 \\
 & none & 0.3227 & \textbf{-66.96} \\
\bottomrule
\end{tabular}
\caption{Macro F1 scores and percentage delta values for select models provided by Qwen}
\label{tab:complete_ablation_results_qwen}
\end{table}

\subsection{Prompt Ablation Study Figures By Provider}

\begin{figure}[!htbp]
  \includegraphics[width=\columnwidth]{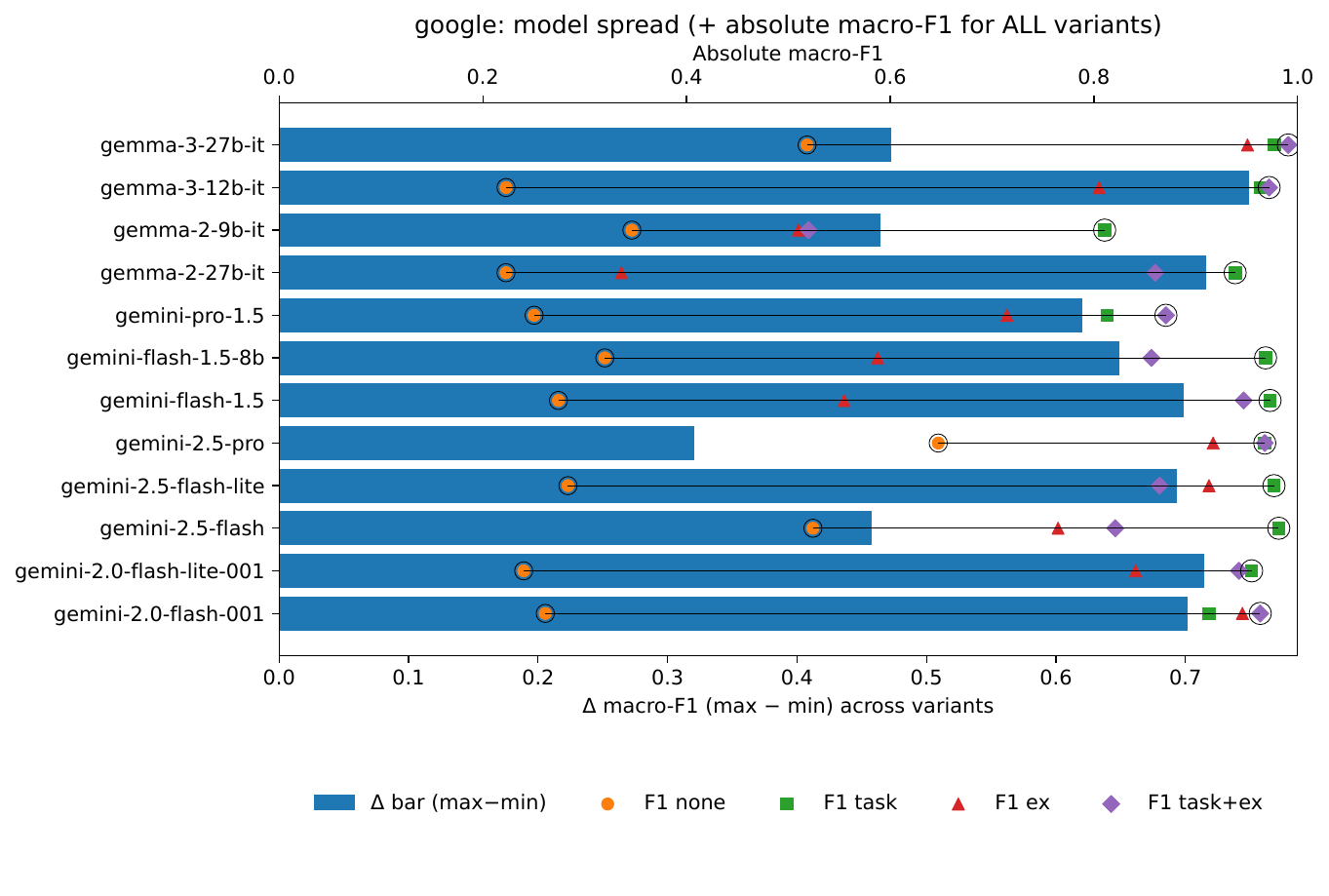}
  \caption{Distribution of macro F1 across prompt versions on gemma model variants. For each model panel, we plot max−min macro F1 (bars) per model and overlay per variant absolute macro F1 (points) on a twin top x-axis, with thin lines showing the min-max span. Models are alphabetized.}
  \label{fig:ablation_spread_google}
\end{figure}

\begin{figure}[!htbp]
  \includegraphics[width=\columnwidth]{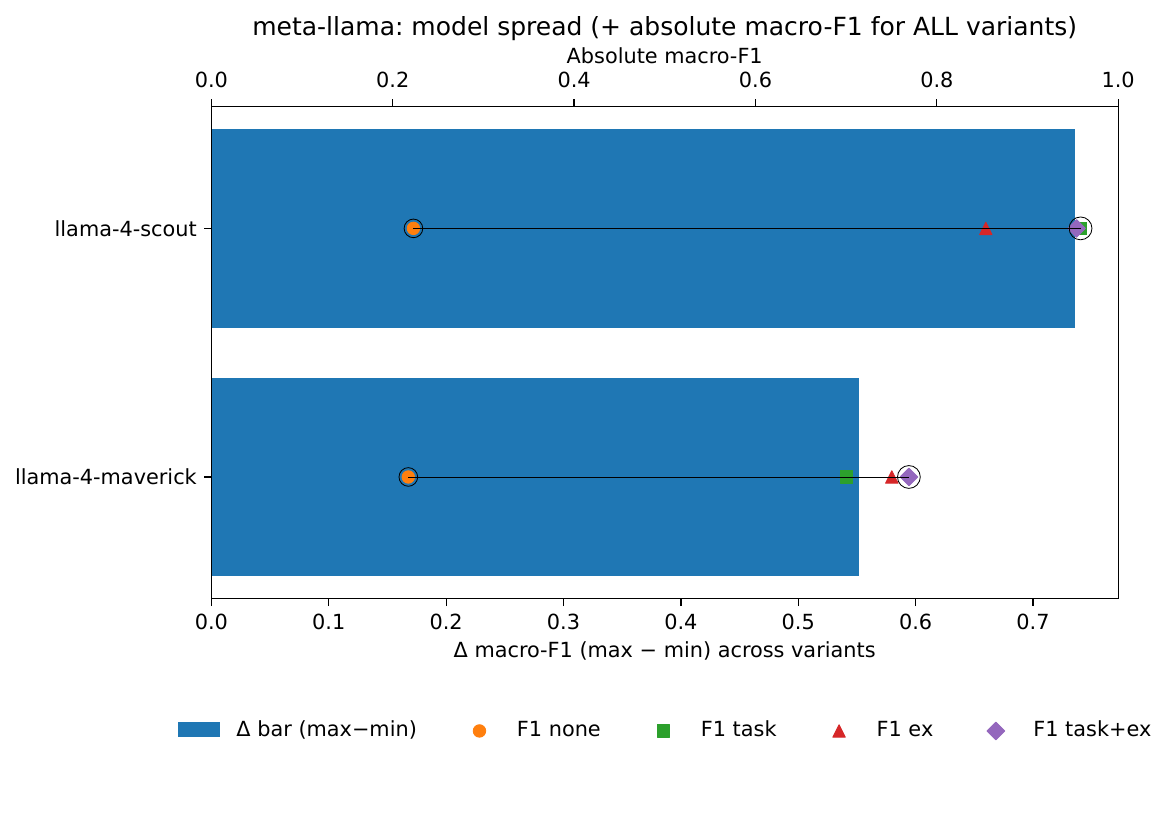}
  \caption{Distribution of macro-F1 across prompt versions on llama model variants . For each model panel, we plot max−min macro-F1 (bars) per model and overlay per-variant absolute macro-F1 (points) on a twin top x-axis, with thin lines showing the min–max span. Models are alphabetized.}
  \label{fig:ablation_spread_llama}
\end{figure}

\begin{figure}[!htbp]
  \includegraphics[width=\columnwidth]{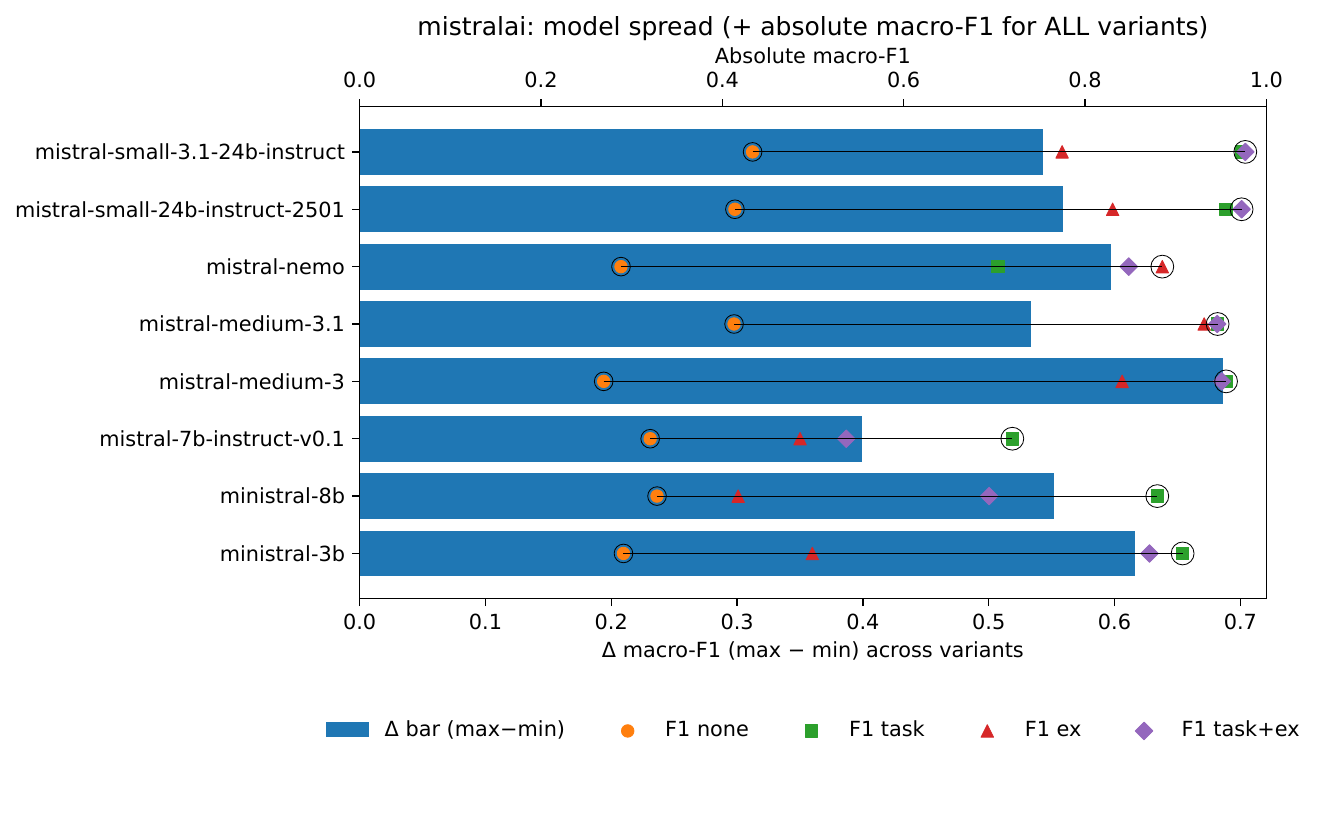}
  \caption{Distribution of macro-F1 across prompt versions on mistral model variants. For each model panel, we plot max−min macro-F1 (bars) per model and overlay per-variant absolute macro-F1 (points) on a twin top x-axis, with thin lines showing the min–max span. Models are alphabetized.}
  \label{fig:ablation_spread_mistral}
\end{figure}

\begin{figure}[!htbp]
  \includegraphics[width=\columnwidth]{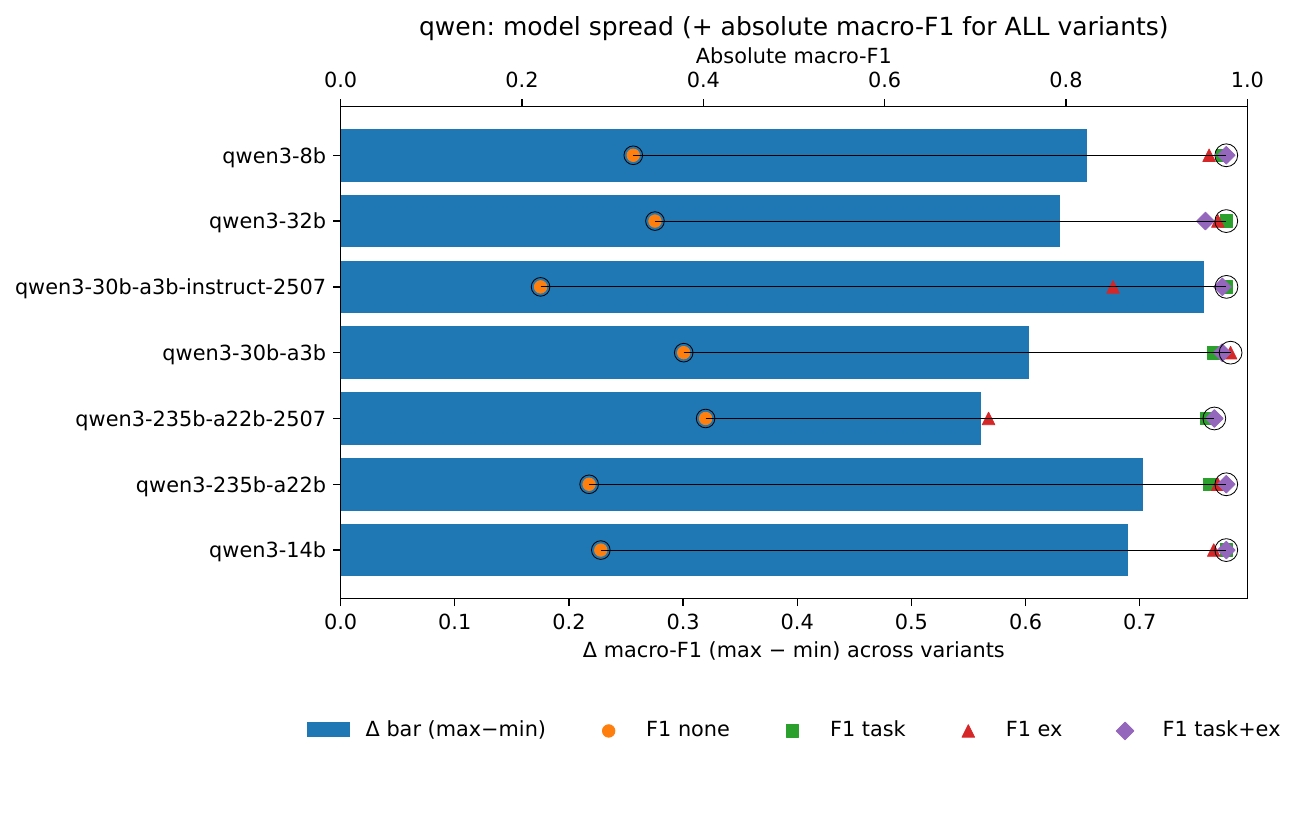}
  \caption{Distribution of macro-F1 across prompt versions on qwen model variants. For each model panel, we plot max−min macro-F1 (bars) per model and overlay per-variant absolute macro-F1 (points) on a twin top x-axis, with thin lines showing the min–max span. Models are alphabetized.}
  \label{fig:ablation_spread_qwen}
\end{figure}

\end{document}